# Robust Load Prediction of Power Network Clusters Based on Cloud-Model-Improved Transformer

Cheng Jiang
*College of Computer and Information Science, Chongqing Normal University,*
Chongqing 401331, China,
384206785@qq.com

Gang Lu
*Energy Strategy and Planning Research Department, State Grid Energy Research Institute Co., Ltd.,*
Beijing 102209, China,
lugang@sgeri.sgcc.com.cn

Xue Ma
*Green Energy Development Research Institute (Qinghai) and the State Grid Qinghai Economic Research Institute,*
Qinghai 810008, China,
mxue2726@qh.sgcc.com.cn

Di Wu
*College of Computer and Information Science, Southwest University,*
Chongqing 400715, China,
wudi.cigit@gmail.com

*Abstract*—Load data from power network clusters indicates economic development in each area, crucial for predicting regional trends and guiding power enterprise decisions. The Transformer model, a leading method for load prediction, faces challenges modeling historical data due to variables like weather, events, festivals, and data volatility. To tackle this, the cloud model's fuzzy feature is utilized to manage uncertainties effectively. Presenting an innovative approach, the Cloud Model Improved Transformer (CMIT) method integrates the Transformer model with the cloud model utilizing the particle swarm optimization algorithm, with the aim of achieving robust and precise power load predictions. Through comparative experiments conducted on 31 real datasets within a power network cluster, it is demonstrated that CMIT significantly surpasses the Transformer model in terms of prediction accuracy, thereby highlighting its effectiveness in enhancing forecasting capabilities within the power network cluster sector.

## I. Introduction

In recent years, as energy consumption rises and the power system grows more complex, power load forecasting has become increasingly challenging. It plays a crucial role in power system scheduling and energy management, significantly impacting the balance between power supply and demand, energy resource optimization, and power grid operation safety and stability. Currently, common techniques for time series forecasting include time series analysis methods such as ARIMA and its variants; neural network models like RNNs; statistical learning methods; nonlinear methods based on machine learning, such as deep learning models, particularly those based on Transformer architecture; ensemble methods that combine multiple models through ensemble learning and stacking to improve prediction accuracy and robustness. These techniques have their own advantages and applicability in various application scenarios, widely used not only in electricity demand forecasting but also in industrial market trend analysis and other analytical domains [1-8]. Consequently, accurately predicting electrical load demand and discovering latent factor data in power big data have emerged as a pressing and intricate issue [9-15].

Recently, the Transformer model has recently become a prominent approach in time series forecasting due to its ability to manage long-term dependencies, leverage parallel computing effectively, handle multi-scale features, offer robust model representation, and provide interpretability. Enhancements to the normalization layer within the Transformer model have yielded improvements in model performance, training efficiency, generalization capacity, and latent factor analysis of big data [16]. These enhancements have bolstered the model's efficacy and resilience in handling intricate data sets and can better find critical information in incomplete and unbalanced data streams [17-19]. Consequently, these strengths have facilitated enhanced performance of advanced learning models in domains like power load forecasting and big data recommendation systems and latent factor analysis [20-23].

**Example**. References [24] and [25] employed distinct approaches to enhance the normalization layer in the Transformer model. Reference [24] introduced column normalization, improving model performance, training efficiency, and generalization across domains. Conversely, reference [25] utilized power normalization (PN) to emphasize partial head characteristics and broaden the channel attention mechanism, addressing conventional normalization layer limitations and enhancing model performance and generalization.

Motivated by the literature cited in references [24] and [25], this study introduces the Cloud Model Improved Transformer (CMIT) model to predict power load, with a focus on enhancing the performance, training efficiency, and generalization capability of the Transformer model. The central concept involves two key aspects: 1) substituting Layer Normalization (LN) with Probability Normalization (PN) based on a cloud model, where the membership algorithm within the cloud model replaces the original normalization layer's method of computing mean and variance in a sequential data order. This adjustment in the normalization layer enhances the model's training stability, convergence speed, and generalization capacity; 2) amalgamating the original normalization layer's Transformer model with the cloud-modelized Transformer model through Particle Swarm Optimization (PSO), thereby boosting the generalization capability of the combined model. Consequently, CMIT preserves all functionalities of the original Transformer model while integrating the resilience of the cloud model and enhancing generalization through PSO [26-31].

This study endeavors to contribute in the following ways:
1) It introduces a novel approach to bolster the resilience of the original Transformer model by integrating insights from the cloud model [32].
2) It presents a technique to substitute the original model's normalization layer with the membership degree derived from the cloud model, facilitating the calculation of mean and variance based on the sequential arrangement of data. This adjustment aims to enhance the stability and consistency of the normalization layer with respect to the data.
3) Finally, combining the optimized Transformer model with the cloud model into a unified CMIT model, using particle swarm optimization, enhances both performance and training stability.

Sec II introduces related work. Sec III states preliminaries. Sec IV presents a CMIT model. Sec V reveals experimental results. Sec VI gives some discussions.

## II. RELATED WORK

The Transformer model, following its notable success in natural language processing tasks, has emerged as a potent instrument for analyzing and predicting electricity time series data. Its utilization in this field has resulted in notable progress in power load demand prediction , smart grid energy consumption optimization [33], and grid stability enhancement through voltage forecasting. Furthermore, the Transformer model has significantly contributed to renewable energy integration and forecasting, facilitating the efficient utilization of solar and wind power sources. Beyond these primary applications, the Transformer model has been applied in anomaly detection for power grid security, fault diagnosis in power systems [34-35], and demand response management for energy efficiency. These diverse applications underscore the Transformer model's versatility, security, and effectiveness in handling intricate electricity time series data, establishing it as a pivotal technology for propelling the energy sector towards a more sustainable.

Numerous enhancements can be made to the Transformer model, such as attention enhancement, context awareness, multi-task learning, latent factorization learning, uncertainty awareness, graph enhancement, depth regularization, sparsity constraints, diversity-driven mechanisms, and awareness of data characteristics [36-38]. Although the Transformer model demonstrates diverse architectures and functionalities, none of them integrate cloud modeling techniques to enhance its performance. We've developed a Cloud-Model-Improved Transformer (CMIT) model that incorporates cloud model principles to boost performance. Unlike traditional Transformers, CMIT uses a prediction-sampling-based multi-model structure inspired by Particle swarm principles. This structure systematically enhances information density in specific time series datasets, improving the Transformer's ability to learn representations. In time series forecasting tasks, CMIT outperforms conventional Transformers in terms of generalization.

## III. Preliminaries

The Transformer model, a deep learning model built on the attention mechanism, outperforms traditional recurrent neural networks (RNNs) [39] and long short-term memory networks (LSTMs) in handling long-distance dependencies, moreover, deep learning can also be applied in time series forecasting, recommender systems, latent factor analysis and cluster analysis [40-52]. The core concept of the cloud model posits that any event or concept can be represented as a "cloud," encapsulating various possibilities and their distributions through membership functions. By merging and manipulating multiple "clouds," uncertain information can be accurately described and processed. Particle Swarm Optimization (PSO) is an optimization algorithm grounded in swarm intelligence, drawing inspiration from the collective behavior of bird flocks or fish schools, and It can also be used for latent factor analysis [53-71]. Initially proposed in the 1990s by American social psychologists Kennedy and Eberhart for solving optimization problems.

***Definition 1 (Cloud model).*** Let $U$ be the domain, $C$ be a qualitative concept on the domain $U$, and $C$ is represented by three numerical features: *Ex*, *En*, and *He*. A precise numerical value $x \in U$ is a realization of $C$ as a one-time normal random event, and $x$ follows a normal distribution with an expectation of *Ex* and a variance of $En'^2$, denoted as $x \sim N(Ex, En'^2)$, where $En'^2$ is a realization of a normal distribution with an expectation of *Ex* and a variance of $He'^2$, denoted as $En' \sim N(En, He'^2)$. The certainty degree $u(x)$ of the numerical value $x$ belonging to $C$ satisfies:

$$u(x) = e^{\frac{-(x-Ex)^2}{2(En')^2}}. \quad (1)$$

The reverse cloud generator's primary function is to calculate the sample mean $X$ and variance $S^2$, and compute the formulas for *Ex*, *En*, and *He* as follows:

$$Ex = X, \quad (2)$$

$$En = \sqrt{\frac{\pi}{2}} \frac{\sum_{i=1}^{n} |x_i - Ex|}{n}, \quad (3)$$

$$He = \sqrt{|S^2 - En^2|}, \quad (4)$$

where $x$, *Ex*, *En*, *He*, and $u(x)$ are generated in the forward cloud and reverse cloud generators of the cloud model. The primary function of the forward cloud generator is to generate a normal random number $En'$ with *En* as the expectation and $He'^2$ as the variance. It also generates a normal random number $x$ with *Ex* as the expectation and $En'^2$ as the variance, and finally generates $u(x)$.

***Definition 2 (Particle Swarm Optimization).*** The particle swarm optimization randomly generates $q$ particles, each of which contains two vectors for position and velocity. The position vector comprises the weight coefficient vector $W_i = (w_{i,1},...,w_{i,n+1})$, where $w_{i,j} \in [0,1], 1 \leq j \leq n+1$, and $w_{i,1}+w_{i,2}+...+w_{i,n+1}=1$. The velocity vector $V_i=(v_{i,1},...,v_{i,n+1})$ ($v_{i,j} \in [-1,1], 1 \leq j \leq n+1$) is used to control the update speed of $W_i$. The update formulas for vectors $V_i$ and $W_i$ at the t-th iteration are as follows:

$$V_{i,t} = \omega \times V_{i,t-1} + C_1 \times r_1 \times (P_{best,i} - W_{i,t-1}) + C_2 \times r_2 \times (P_{gbest} - W_{i,t-1}), \quad (5)$$

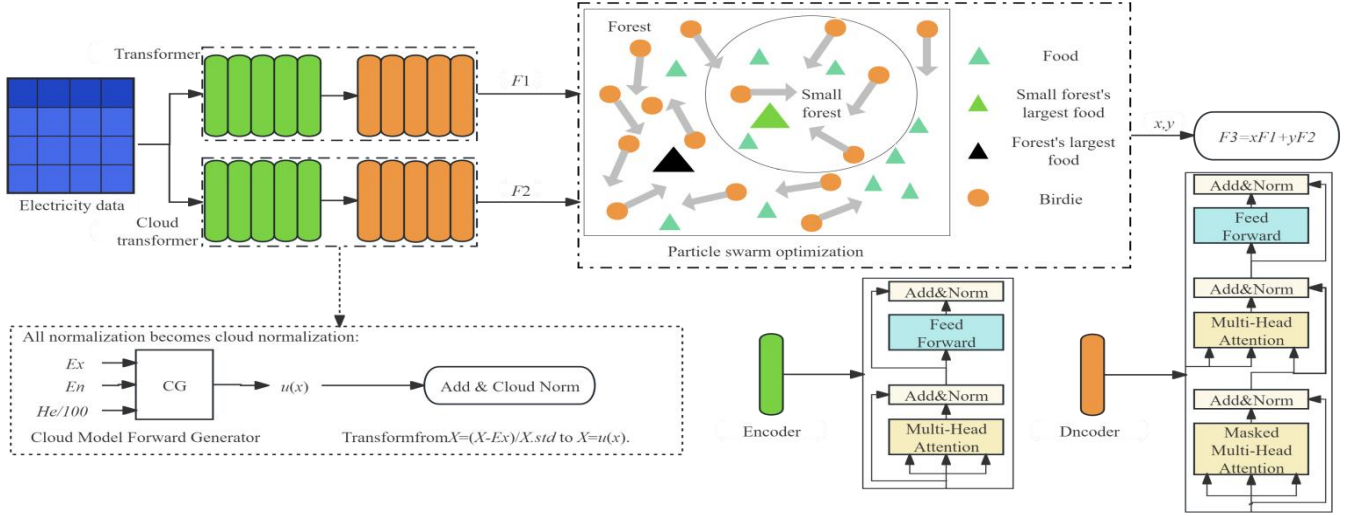

Fig. 1. The structure of CMIT model.

$$W_{i,t} = W_{i,t-1} + V_{i,t}, \qquad (6)$$

where $\omega$ is the inertia weight coefficient, $C_1$ and $C_2$ are the cognitive and social coefficients (also known as acceleration constants) respectively, $r_1$ and $r_2$ are two standard normal distribution random numbers, $P_{best,i}$ is the best position vector of particle i, $P_{gbest}$ is the best position vector of all particle, and $P_{best,i}$ and $P_{gbest}$ are initialized by $W_i$. In each iteration, the optimization objective function of particle i is calculated as follows:

$$f = avg\_error_t = \|Y - X \bullet W_i\|. \qquad (7)$$

Firstly, update $P_{best,i}$ and $P_{gbest}$ based on the value of the optimization objective function: if the optimization objective function value of particle i in this iteration is less than the optimization objective function value of $P_{best,i}$, then let $P_{best,i} = W_i$; if the optimization objective function value of $P_{best,i}$ is less than the optimization objective function value of $P_{gbest}$, then let $P_{gbest}=P_{best,i}$. Then, update $V_i$ and $W_i$ according to formulas (5) and (6). When the predefined maximum number of iterations is reached, the algorithm terminates, and $P_{gbest}$ at this point is the final optimal weight vector $W$.

IV. PROPOSED CMIT MODEL

*A. Structure of CMIT*

Following the flowchart depicted in Figure 1, the design principle of the proposed Cloud-Model-Improved Transformer (CMIT) model involves initially substituting the normalization layer in the original Transformer model with cloud model membership degrees. These degrees are computed based on the sequential arrangement of data to determine the mean and variance. Subsequently, the power data is fed into both the original Transformer model and the Transformer model normalized by clouds. The resulting dataset is then fed into the particle swarm optimization algorithm, which analyzes the latent factors and generates the required prediction data [72-91]. The operational steps are as follows:

1) Input the forward generator and reverse generator of the cloud model;
2) Replace the method of calculating mean and variance in the original Transformer model's normalization layer, based on data sequence, with computing membership degrees in the cloud model to derive method *F*2;
3) Feed the data into Transformer for training and obtaining method *F*1;
4) Train the data in the Transformer model normalized by clouds to acquire method *F*2;
5) Introduce *F*1 and model *F*2 to the particle swarm optimization algorithm;
6) Provide the parameters of the particle swarm optimization algorithm, including the number of particles *x* and iterations *y*;

7) Generate method *F*3 through the particle swarm optimization algorithm.
8) Input the power data into method *F*3;
9) Output the predicted data.

*B. Obtaining the cloud-normalized Transformer model*

To explain how to obtain the cloud-normalized Transformer model, we describe the process of inputting data into the normalization layer in the general case as $X \in \{x_1, ... x_n\}$. While equations (2), (3), and (4) remain unchanged, equation (1) is expanded as follows:

$$x_i = u(x_i) = e^{\frac{-(x_i - Ex)^2}{2(random(loc=En, scale=He/100))^2}}, \quad (8)$$

where $u(x_i)$ represents the membership degree of the *i*-th data in this layer, *Ex* denotes the mean of the layer's data, *En* denotes the entropy of the layer's data, *He* denotes the hyper-entropy of the layer's data, and *random(loc,scale)* represents the generation of a normal random number with mean *loc* and standard deviation *scale*.

*C. Training model C using the particle swarm algorithm*

To explain how to construct a particle swarm training set to learn the optimal weights for method *F*3, we define *T*=(*X*,*Y*) as the training set, where *X* is the predicted values corresponding to the preceding *p* time steps for all particle layers at the current time, and *Y* is the corresponding observed values. In other words, *X* is a 2*p*-dimensional matrix, and *Y* is a *p*-dimensional vector:

$$X = \begin{bmatrix} A_1 & B_1 \\ \vdots & \vdots \\ A_i & B_i \\ \vdots & \vdots \\ A_p & B_p \end{bmatrix}, Y = \begin{bmatrix} R_1 \\ \vdots \\ R_i \\ \vdots \\ R_p \end{bmatrix}, \quad (9)$$

where $A_i$ represents the predicted value at time *i* by the original Transformer model, $B_i$ represents the predicted value at time *i* by the cloud-normalized Transformer model, and $R_i$ represents the true value at time *i*. Finally, we continuously train method *F*3 using equations (5), (6), (7):

$$F3 = (F1 + F2)W, \quad (10)$$

where *F*1 represents the original Transformer path, *F*2 represents the cloud-normalized Transformer path, and *W* represents the optimal weight matrix.

## V. EXPERIMENTS

In the conducted experiments, we endeavor to address the following research questions (RQs):

**RQ. 1.** To what extent does substituting the original Transformer model's normalization layer with membership degrees in the cloud model enhance the performance of the Transformer model within a defined scope?

**RQ. 2.** Does the Cloud-Model-Improved Transformer (CMIT) model outperform the original Transformer model in predicting power load?

*A. General Settings*

**Datasets.** The experimental dataset comprises the daily power load data from 31 power network clusters in Haimen City, China, spanning from January 1, 2021, to February 28, 2023. Each power network cluster represents a distinct street and commercial consumption scenario. We partitioned the dataset into training, validation, and testing sets as follows: January 1, 2021, to December 31, 2021, and March 1, 2022, to December 31, 2022, for training; January 1, 2022, to February 28, 2022, for validation; and January 1, 2023, to February 28, 2023, for testing.

**Evaluation Protocol.** In various electric power system applications, precise forecasting of future power loads holds paramount importance. Hence, power load forecasting stands as a pivotal tool for understanding load system characteristics. Mean Absolute Percentage Error (MAPE) commonly serves as the metric for assessing prediction accuracy, calculated as follows:

$$MAPE = \frac{100\%}{n} \sum_{i=1}^{n} \left| \frac{\hat{y}_i - y_i}{y_i} \right|,$$

where $|\cdot|_{abs}$ indicates the absolute value.

**Baselines.** The evaluation of the CMIT model is outlined in Table I, detailing the experimental usage model. We performed experiments utilizing a substantial dataset on three variants: the original Transformer model, the cloud-normalized Transformer model, and the CMIT model. This process generated three distinct sets of experimental data, which were subsequently compared using the Mean Absolute Percentage Error (MAPE).

TABLE I. THE DESCRIPTION OF THE MODELS AND ALGORITHMS USED.

| Model/Algorithm | Description |
|---|---|
| Transformer model | Utilizing in time series analysis for efficient temporal dependency capture via self-attention mechanisms, enabling effective sequence modeling and forecasting. |
| Cloud Transformer model | A model that replaces the model of the original Transformer model normalization layer with the membership degree in the cloud model. |
| Particle swarm optimization algorithm | It is commonly employed to iteratively enhance a population of candidate solutions to optimization problems based on predefined criteria. |
| CMIT model | A model that combines the original Transformer model with the cloud Transformer model using the particle swarm optimization algorithm. |

**Implementation Details.** We standardized parameters across CMIT, Transformer, and Cloud Transformer models, including optimizer (Adam), learning rate (lr=0.001), regularization coefficients, and training iterations. This uniform approach ensures optimal prediction accuracy for each model. Experiments were conducted on a computer server with a 2.1 GHz E5-2620 CPU, 32 cores, and 256 GB RAM, with results averaged across multiple tests.

TABLE II. THE COMPARISON RESULTS ON MAPEs INCLUDING FRIEDMAN TEST AND WIN/LOSS COUNTS.

| Dataset | Transformer | Cloud Transformer | CMIT |
|---|---|---|---|
| D1 | 29.13 | 30.79 | **15.11** |
| D2 | 34.34 | 35.37 | **34.24** |
| D3 | 68.32 | 68.00 | **60.05** |
| D4 | 10.96 | 12.66 | **10.35** |
| D5 | 46.96 | 50.48 | **46.92** |
| D6 | 13.32 | 13.56 | **13.16** |
| D7 | 37.79 | 37.06 | **36.17** |
| D8 | 10.31 | 10.74 | **8.07** |
| D9 | 17.21 | 17.90 | **13.42** |
| D10 | 21.63 | **18.77** | 21.65 |
| D11 | 16.24 | 12.28 | **12.17** |
| D12 | 18.37 | 20.28 | **11.60** |
| D13 | 18.84 | 21.49 | **17.84** |
| D14 | 30.41 | 35.40 | **30.34** |
| D15 | 17.82 | 24.41 | **17.71** |
| D16    | 16.83    | **11.24**| 12.05    |
| D17    | 16.01    | **12.21**| 13.45    |
| D18    | 16.69    | 18.53    | **15.08**|
| D18    | 7.98     | 9.82     | **7.94** |
| D20    | 18.29    | 18.60    | **18.19**|
| D21    | 49.98    | 47.69    | **45.34**|
| D22    | 37.31    | **33.18**| 36.95    |
| D23    | 26.22    | 22.02    | **20.32**|
| D24    | **17.91**| 22.69    | 18.28    |
| D25    | 51.52    | 50.87    | **46.79**|
| D26    | 14.41    | 11.85    | **11.65**|
| D27    | 17.63    | 19.42    | **17.51**|
| D28    | 24.79    | **15.37**| 19.24    |
| D29    | 22.79    | 16.68    | **16.20**|
| D30    | 18.57    | 18.31    | **17.45**|
| D31    | 17.89    | 19.16    | **17.11**|
| Mean-MAPE | 24.72 | 24.41    | 22.01    |
| Win/Loss  | 1/30  | 5/26     | 25/6     |
| F-rank    | 2.35  | 2.32     | 1.32     |
| *p*-value | 0.0000026 | 0.00061 | \      |

* The 'Win' means that the model is the best of the three models, and the 'Loss' means that the model is not the best of the three models.

*A. Performance Comparison (RQ.1)*

Statistical analysis was conducted on the comparison results documented in Table II to enhance the understanding of these MAPE comparisons. The win-loss count between the Cloud Transformer model and the Transformer model is displayed in the final two rows of Table III. Additionally, a Friedman test was performed to assess the models' performance across various datasets and compare them based on MAPE. The statistical outcomes of the Friedman test are presented in the last row of Table III. Table II reveals that while the Friedman results for the Cloud Transformer model and the original Transformer model align, the Cloud Transformer model demonstrates slightly higher average MAPE values and Win values than the original Transformer model, suggesting superior performance within a specific range. These findings indicate that replacing the normalization layer in the original Transformer model with membership degrees in the Cloud model has indeed enhanced the Transformer model's performance to some extent.

*B. Performance comparison of the CMIT model (RQ.2)*

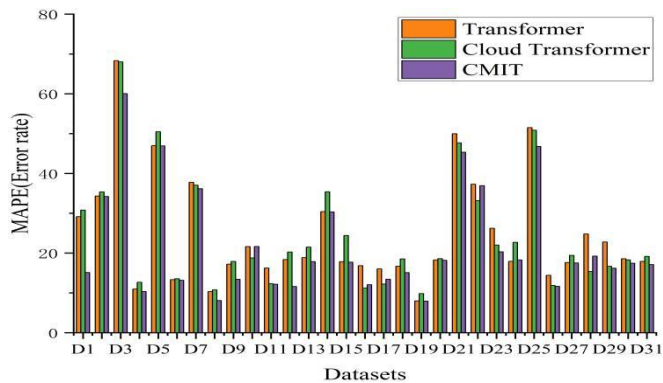

Fig. 2. MAPE variations of different models across all datasets during the training process.

TABLE III. THE WILCOXON SIGNED-RANKS TEST RESULTS ON MAPE OF TABLES II.

| Comparison | R+ | R- | p-value* |
|---|---|---|---|
| CMIT vs. Transformer | 481 | 15 | **0.0000026** |
| CMIT vs. Cloud Transformer | 413.5 | 82.5 | **0.00061** |

* The accepted hypotheses with a significance level of 0.05 are highlighted.

Based on the analysis of the experimental data in Table II, the bar chart depicted in Figure 2, and the *p*-value table in Table III, the following observations can be made:

1) From the statistical outcomes presented in Table II, it is evident that among the 31 power network clusters datasets, the CMIT model performs exceptionally well. Specifically, out of the total datasets, only one exhibits a slightly higher MAPE value compared to the original Transformer model, five datasets show higher MAPE values than the Cloud Normalization Layer Transformer model, and two datasets have MAPE values equivalent to the original Transformer model. Notably, the CMIT model outperforms the remaining 25 datasets. Additionally, the average MAPE across all 31 datasets for the CMIT model is significantly lower than both the original Transformer model and the Cloud Normalization Layer Transformer model.

2) Upon examination of the bar chart in Figure 2, it is apparent that the majority of bars representing the CMIT model are the lowest, indicating superior predictive performance across most of the 31 datasets.

3) Furthermore, the Wilcoxon test results in Table III reveal that the *p*-value of CMIT relative to the original Transformer model is 0.0000026, and relative to the Cloud Transformer model is 0.00061, both significantly below the 0.05 threshold. This suggests a substantial difference in performance.

In conclusion, the integration of the cloud model (CMIT) has notably enhanced the performance and generalization capabilities of the original Transformer model in time series electricity load forecasting.

## VI. CONCLUSION

This study adheres to the principles of time series analysis forecasting and introduces the Cloud Model Improved Transformer model (CMIT) for power network clusters load prediction. CMIT amalgamates the strengths of the cloud model and Transformer model while integrating the particle swarm optimization algorithm to enhance its generalization capacity towards the data. Experimental validation of the proposed model was conducted using 31 datasets representing diverse electricity consumption scenarios across different regions. The findings reveal that: 1) The Cloud Normalization Layer Transformer model demonstrates a slight performance edge over the original Transformer model within a specific range, and 2) CMIT surpasses both the original Transformer model and the Cloud Normalization Layer Transformer model in the accurate prediction of power network clusters load data.